\documentclass[10pt,twocolumn,letterpaper]{article}
\usepackage[final]{wacv}
\usepackage{graphicx}
\usepackage{amsmath}
\usepackage{amssymb}
\usepackage{booktabs}
\usepackage[colorinlistoftodos]{todonotes}
\usepackage{multirow}
\usepackage[pagebackref,breaklinks,colorlinks]{hyperref}
\usepackage[capitalize]{cleveref}
\usepackage{adjustbox}
\crefname{section}{Sec.}{Secs.}
\Crefname{section}{Section}{Sections}
\Crefname{table}{Table}{Tables}
\crefname{table}{Tab.}{Tabs.}

\def\wacvPaperID{2550} \def\confName{WACV}
\def\confYear{2025}

\begin{document}

\title{Diffusion-based Visual Anagram as Multi-task Learning}

\author{
Zhiyuan Xu\textsuperscript{*1,2},
Yinhe Chen\textsuperscript{*1,3},
Huan-ang Gao\textsuperscript{1},
Weiyan Zhao\textsuperscript{1,4},
Guiyu Zhang\textsuperscript{1},
Hao Zhao\textsuperscript{\textdagger 1}
\\
\textsuperscript{1}Institute for AI Industry Research (AIR), Tsinghua University\\
\textsuperscript{2}University of Chinese Academy of Sciences\\
\textsuperscript{3}Central South University,
\textsuperscript{4}Beijing Institute of Technology
}

\maketitle
\setcounter{footnote}{1}
\footnotetext{\textsuperscript{*}Equal Contribution. \textsuperscript{\textdagger}Corresponding Author.}
\setcounter{footnote}{2}
\footnotetext{Code: \url{https://github.com/Pixtella/Anagram-MTL}}

\begin{abstract}
\vspace{-2mm}
Visual anagrams are images that change appearance upon transformation, like flipping or rotation. 
With the advent of diffusion models, generating such optical illusions can be achieved by averaging noise across multiple views during the reverse denoising process.
However, we observe two critical failure modes in this approach:
(i) concept segregation, where concepts in different views are independently generated,
which can not be considered a true anagram,
and (ii) concept domination, where certain concepts overpower others.
In this work, we cast the visual anagram generation problem in a multi-task learning setting, where different viewpoint prompts are analogous to different tasks, and derive denoising trajectories that align well across tasks simultaneously.
At the core of our designed framework are two newly introduced techniques, where (i) an anti-segregation optimization strategy that promotes overlap in cross-attention maps between different concepts, and (ii) a noise vector balancing method that adaptively adjusts the influence of different tasks.
Additionally, we observe that directly averaging noise predictions yields suboptimal performance because statistical properties may not be preserved, prompting us to derive a noise variance rectification method.
Extensive qualitative and quantitative experiments demonstrate our method's superior ability to generate visual anagrams spanning diverse concepts.
\end{abstract}

\vspace{-3mm}
\section{Introduction}
\vspace{-1mm}
\label{sec:intro}

\begin{figure}
    \centering
    \includegraphics[width=0.8\linewidth]{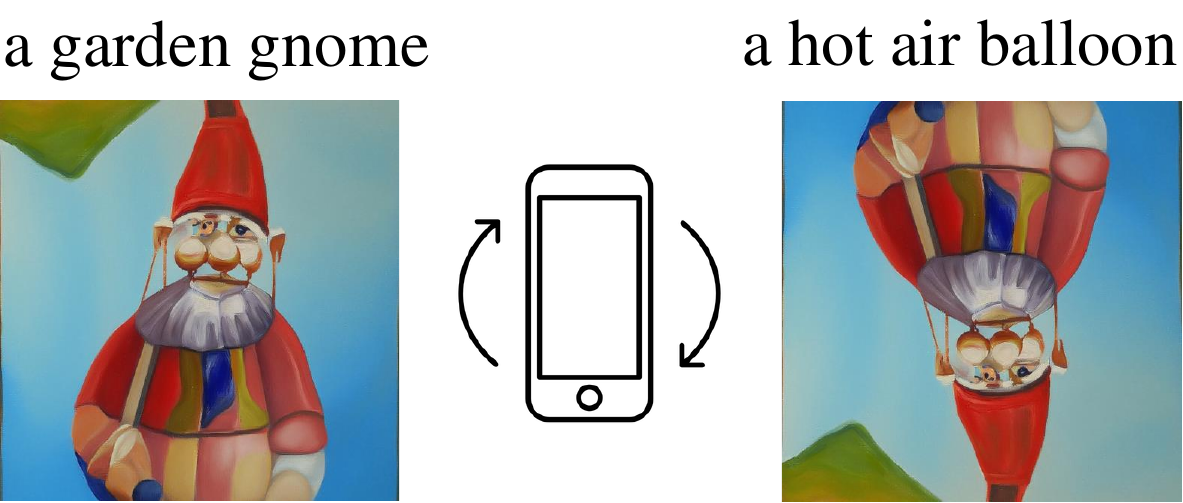}
    \vspace{-1mm}
    \caption{\textbf{Visual Anagrams.} We show an example of visual anagrams, which can be perceived as a garden gnome or a hot air balloon depending on the orientation of the image.}
    \label{fig:gd1}
    \vspace{-3mm}
\end{figure}

\begin{figure}
    \centering
    \begin{subfigure}{.47\linewidth}
        \includegraphics[width=\linewidth]{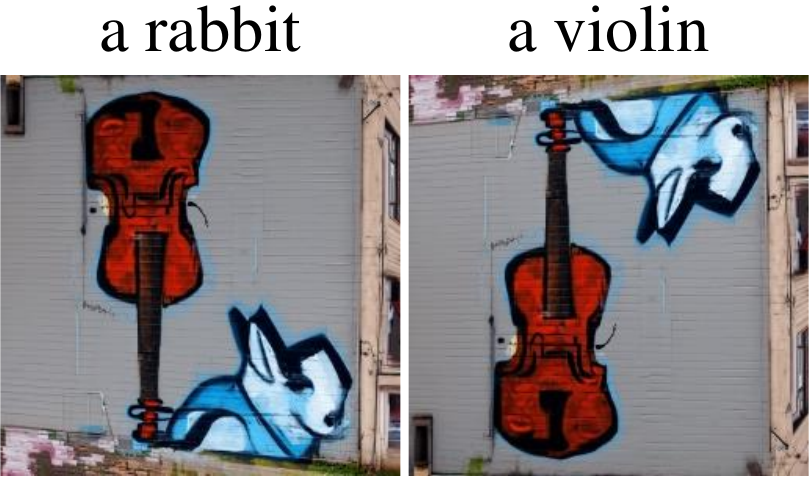}
        \caption{a street art of ...}
        \vspace{-1mm}
        \label{fig:fail1}
    \end{subfigure}
    \hskip0.3em
    \begin{subfigure}{.47\linewidth}
        \includegraphics[width=\linewidth]{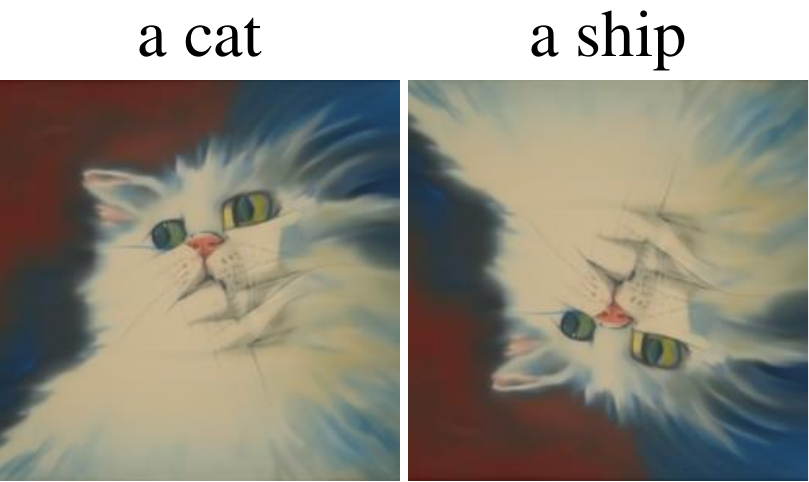}
        \caption{a painting of ...}
        \vspace{-1mm}
        \label{fig:fail2}
    \end{subfigure}
    \caption{\textbf{Failure Cases.} We show two common failure cases of \cite{geng2024visual}: concept segregation (left) and domination (right). 
    \label{fig:fail}
    }
    \vspace{-5mm}
\end{figure}

\textbf{Visual anagrams} are images whose appearances change when subjected to transformations such as rotation or flipping \cite{geng2024visual,burgert_diffusion_2024}. 
For instance, as illustrated in \cref{fig:gd1}, an image may resemble a garden gnome when viewed head-on but transform into a hot air balloon when observed upside-down.
The study of visual anagrams involves understanding how geometric transformations manipulate and redefine visual perception, making this field promising for applications in art, design, and cognitive science \cite{boring1930new,ferino2017arcimboldo,nicholls_perception_2018}.

\textbf{Diffusion.} With the advent of diffusion models \cite{sohl-dicksteindeepunsupervisedlearning_2015, hodenoisingdiffusionprobabilistic_2020,dhariwaldiffusionmodelsbeat_2021,
songscorebasedgenerativemodeling_2020, zhang2024ctrl}, generating such visual anagrams has become significantly easier by simply averaging noise predictions across different views of various prompts. In the reverse denoising process of diffusion models, the image layout is formed during the early denoising timestamps while the colors and details are generated in later timestamps. This allows us to establish a layout of noised images that matches the different prompts before adding details. This aligns with the ideas that generative models and human perception may process optical illusions similarly \cite{gomez-villaconvolutionalneuralnetworks_2019, jainiintriguingpropertiesgenerative_2024, ngoclipfooledoptical_2023a}.

\textbf{Problems.} Despite the simplicity of this method, our careful investigation has identified two critical issues. First, there is the problem of concept segregation, where different concepts are synthesized independently (as illustrated in \cref{fig:fail1}, where the rabbit and the violin appear as separate objects within a single image), which does not constitute a true visual anagram.
Secondly, there is the issue of concept domination, where some concepts overpower others in the generated images (as shown in \cref{fig:fail2}, where the cat is clearly visible but the ship is not as distinct).

\textbf{Solutions.} 
To address these two issues, we draw inspiration from multi-task learning, aiming to derive denoising trajectories that enhance commonalities across multiple generation tasks while adaptively alleviating their conflicts.
At the core of our solution lie two new techniques:
\textbf{(I) Anti-Segregation Optimization}: By treating the attention map of spatial queries against the subjects of different prompts as encoding weights for different tasks in the intermediate images being denoised, we effectively prevent the segregation of concepts by promoting the intersection of these attention score maps across different views.
\textbf{(II) Noise Vector Balancing}: Noise predicted from different views can be seen as optimizing directions (or gradients) for different tasks in the denoising trajectory. We adaptively balance the noise vectors across tasks by reweighting them using a task completion measure.
\textbf{Additionally}, we recognize the importance of preserving variance in the diffusion denoising process. To this end, we modulate the combined noise predictions with estimated noise correlation information to maintain key statistical properties.

In summary, our contributions are as follows:
\begin{itemize}
    \vspace{-2mm}
    \item \textbf{Integration with Multi-Task Learning}: We link the visual anagram generation problem to multi-concept generation and propose the use of multi-task learning techniques to address it effectively.\vspace{-2mm}
    \item \textbf{Solution to Concept Segregation and Domination}: We identify critical issues, such as concept segregation and domination in generated images, and propose two effective methods to resolve them.\vspace{-2mm}
    \item \textbf{Comprehensive Evaluation}: We present extensive quantitative and qualitative results to demonstrate the efficacy and flexibility of our proposed method compared to previous satet-of-the-art\cite{geng2024visual}.
\end{itemize}

\begin{figure*}[t]
    \centering
    \includegraphics[width=\textwidth]{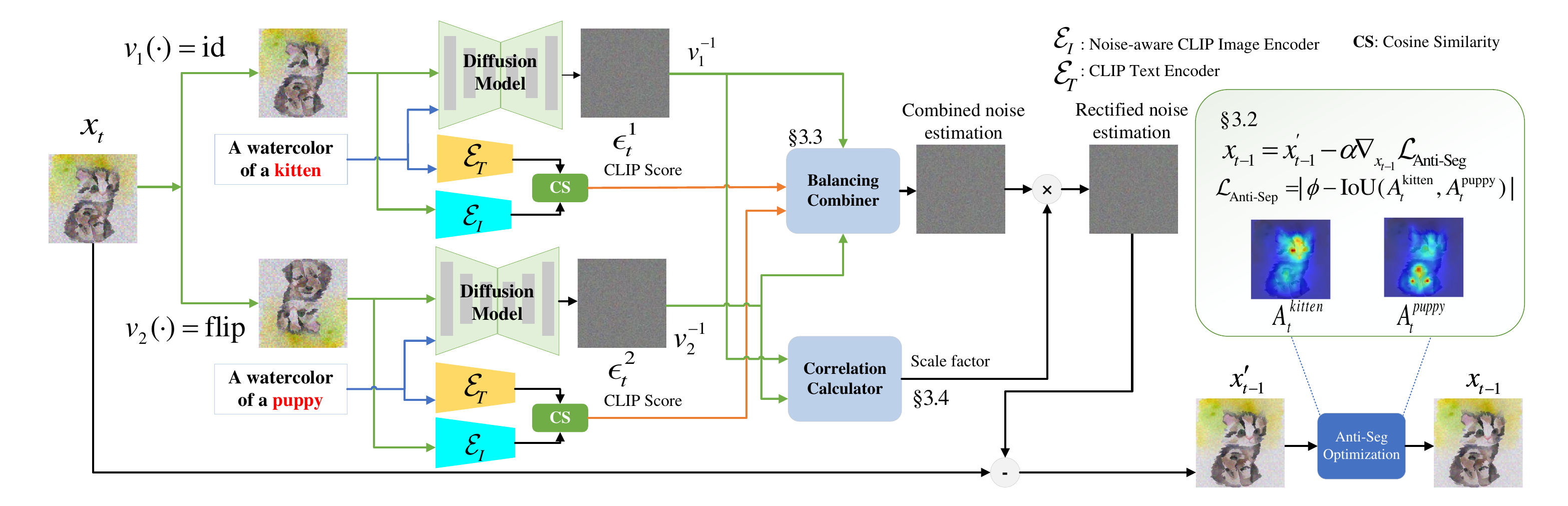}
    \vspace{-7mm}
    \caption{\textbf{Method overview.} During each denoising step, the intermediate image $x_t$ first passes through the diffusion model together with the corresponding text prompt under each view, and also through a noise-aware CLIP model which measures the degree of task completion for each view. \textbf{(1) Noise Vector Balancing:} Predicted noise vectors are reweighted based on the degree of task completion before being combined, see \cref{sec:noise-balancing}. \textbf{(2) Noise Variance Rectification:} Combined noise vectors are rectified by applying a scale factor calculated based on estimated correlation coefficients, which is detailed in \cref{sec:noise-unitization}. \textbf{(3) Anti-Segregation Optimization:} The denoised image $x_{t-1}'$ is modulated to encourage intersection of attention maps of different views with an inference-time loss term before being passed to the next denoising step, refer to \cref{sec:anti-se}.}
    \vspace{-5mm}
    \label{fig:method_overview}
\end{figure*}

\vspace{-3mm}
\section{Related Work}
\vspace{-1mm}

\textbf{Computational Optical Illusions.} 
Optical illusions and computational arts hold great potential for advancing our understanding of human and machine perception \cite{gomez-villaconvolutionalneuralnetworks_2019,hertzmannvisualindeterminacygan_2020,jainiintriguingpropertiesgenerative_2024,wangquantifyingambiguitiesartistic_2020, ngoclipfooledoptical_2023a,lu2017decoder,lu_closed-form_2019}.
A long line of research has focused on generating such illusions by leveraging specific characteristics of human vision. 
For instance, creating the illusion of continuous uni-direction motion can be achieved by locally applying a filter with a continuously shifting phase, based on the observation that local phase shifts are interpreted as global movement \cite{freemanmotionmovement_1991}.
Hybrid images, which change appearance depending on viewing distance, can be generated by exploiting the multiscale nature of human perception, blending high frequencies from one image with low frequencies from another \cite{hybridimages}.
Camouflage images, which contain hidden figures that remain imperceptible or unnoticed at first glance, can be created by removing the original subtle texture details of the hidden figures and replacing them with those of the surrounding prominent image, as human perception operates on a two-phase basis: feature search and conjunction search \cite{chucamouflageimages_2010, guoganmouflage3dobject_2023,owenscamouflagingobjectmany_2014}
Perceptual puzzles involving color constancy, size constancy, and face perception can be generated by finding adversarial examples for principled models of human perception, where vision is explicitly modeled as Bayesian inference \cite{chandradesigningperceptualpuzzles_2022}.
While this line of research shows great promise, the dependence on explicitly identifying characteristics within human vision models can be cumbersome. And this challenge can be alleviated by leveraging the data priors learned by diffusion models \cite{hodenoisingdiffusionprobabilistic_2020,gengvisualanagramsgenerating_2023}.

\textbf{Diffusion-based Visual Anagram generation.}
Diffusion models \cite{dhariwaldiffusionmodelsbeat_2021,graikosdiffusionmodelsplugplay_2023,hodenoisingdiffusionprobabilistic_2020,rameshzeroshottextimagegeneration_2021,rombachhighresolutionimagesynthesis_2021,sahariaphotorealistictextimagediffusion_2022,sohl-dicksteindeepunsupervisedlearning_2015,songdenoisingdiffusionimplicit_2020,songscorebasedgenerativemodeling_2020, li2024fairdiff, gao2024scp, chen2024ultraman} are a powerful class of generative models that gradually transform a noise-distributed sample into one from a target data distribution by following a reverse denoising trajectory. 
In this process, the models estimate and remove noise over time, conditioned on both a noisy image and a specific timestep. This is further enhanced by incorporating an embedding of a text prompt, facilitating text-conditioned image synthesis \cite{rombachhighresolutionimagesynthesis_2021,stabilityaireleases_}.
To generate visual anagrams using diffusion models, Burgert \etal \cite{burgert_diffusion_2024} proposed to use Score Distillation Loss\cite{poole2022dreamfusion}, but this method is computationally expensive and degrades image quality.
Tancik \cite{tancik_tancikillusion-diffusion_2024} demonstrated that visual anagrams can be generated using Stable Diffusion \cite{rombachhighresolutionimagesynthesis_2021} by alternately denoising latents with different prompts under different views.
Geng \etal \cite{geng2024visual} propose to average the noise predictions from different prompts, using a pixel-based diffusion model, and achieve better results in terms of image quality and computational efficiency. Our work builds on the latter approach, which we will refer to as the baseline.

In this paper, we identify two critical issues with this approach: concept segregation and concept domination. We attribute these problems to the failure to explicitly account for the multiple concepts being generated. While compositional multi-concept generation has long been studied \cite{liucompositionalvisualgeneration_2022, cheferattendexciteattentionbasedsemantic_2023, agarwalastartesttimeattention_2023, baoseparateenhancecompositionalfinetuning_2024}, it typically involves creating separate subjects that dominate different regions of an image. 
In contrast, visual anagram generation aims to generate a single object that can be perceived as multiple concepts from different viewpoints. This fundamental difference underscores the unique challenges in generating effective visual anagrams.
To address these issues, we draw an analogy between visual anagram generation and multi-task learning. This analogy enables us to derive denoising trajectories that leverage commonalities between concepts while balancing their differences.

\textbf{Multi-task learning.}
Multi-task learning seeks to enhance the performance and generalization capabilities of models on individual tasks by leveraging shared representations to exploit commonalities and balance differences across tasks\cite{zhang2021survey,NEURIPS2020_634841a6,chen2022cerberus,9682598, zheng2023steps}. 
The primary challenge in this domain is to strike the balance between different tasks, and various techniques have been developed, including loss weighting \cite{guizilinisemanticallyguidedrepresentationlearning_2020, kendallmultitasklearningusing_2018,xincurrentmultitaskoptimization_2022}, gradient normalization \cite{chengradnormgradientnormalization_2018}, gradient dropout \cite{chenjustpicksign_2020}, gradient surgery \cite{yugradientsurgerymultitask_2020}, Nash Bargaining solutions \cite{navonmultitasklearningbargaining_2022}, Pareto-optimal solutions \cite{senermultitasklearningmultiobjective_2019, linparetomultitasklearning_2019,misracrossstitchnetworksmultitask_2016}, gradient alignment \cite{senushkinindependentcomponentalignment_2023}, and curriculum learning \cite{guodynamictaskprioritization_2018a, igarashimultitaskcurriculumlearning_2022}.
In our work, we apply multi-task learning principles to the problem of visual anagram generation. We treat image generation under different views as separate tasks
and develop novel techniques to address task segregation and overpowering issues to yield a denoising trajectory that adaptively balances the differences.

\vspace{-3mm}
\section{Method}
\vspace{-1mm}

Our goal is to generate visual anagrams using off-the-shelf text-to-image diffusion models. Formally, given a set of text prompts $\{y_1, y_2, \ldots, y_N\}$ and corresponding views $\{v_1, v_2, \ldots, v_N\}$, we aim to generate an image $x_0$ that matches each prompt $y_i$ under view $v_i$. This has been shown to be possible by averaging the noise predictions from different prompts \cite{geng2024visual}, which represents previous state-of-the-art, but the resulting images often suffer from concept segregation and concept domination issues as discussed in \cref{fig:fail}. \cref{fig:method_overview} illustrates the overall pipeline of our proposed method. In this section, we first introduce the preliminaries in \cref{sec:pre}, followed by a detailed explanation of our techniques in the following sections.

\vspace{-1mm}
\subsection{Preliminaries}
\vspace{-1mm}
\label{sec:pre}

\textbf{Text-to-Image Diffusion Models.} Diffusion models iteratively denoise a sample $x_T$ initialized from Gaussian noise $\mathcal{N}(0,I)$, to generate a noise-free sample $x_0$ from a learned target distribution. Here, $T$ represents the total number of denoising steps. For text-to-image generation, the model is conditioned on text prompts $y$, and can be formulated as $ \epsilon_{\theta}(x_t, t, y)$, where $x_t$ denotes the intermediate image at time step $t$. The model predicts the noise $\epsilon_t$ at each step, which is then used to compute the next image $x_{t-1}$. In practice, the model could be implemented using the UNet architecture \cite{ronneberger2015u}, with text prompts effectively integrated via cross-attention mechanisms \cite{rombachhighresolutionimagesynthesis_2021}.

\textbf{Text-Image Cross-Attention.} A common method for integrating text prompts into diffusion models is through cross-attention layers within the UNet architecture \cite{rombachhighresolutionimagesynthesis_2021}. In this setup, the projected downsampled image input $x_t^d$ serves as the query, while the projected text embedding $y_e \in \mathbb{R}^{N \times D}$ (where $N$ is the token number and $D$ is the embedding dimension) acts as the key and value. The output of the cross-attention layer is computed using multi-head attention \cite{vaswani2017attention}, where attention maps $A_t \in \mathbb{R}^{r \times r \times N}$ are generated to model the interaction between text prompts and corresponding image regions. Our method in \cref{sec:anti-se} operates on the averaged attention maps from all cross-attention layers at a resolution of $32 \times 32$.

\begin{figure*}
    \centering
    \includegraphics[width=0.95\textwidth]{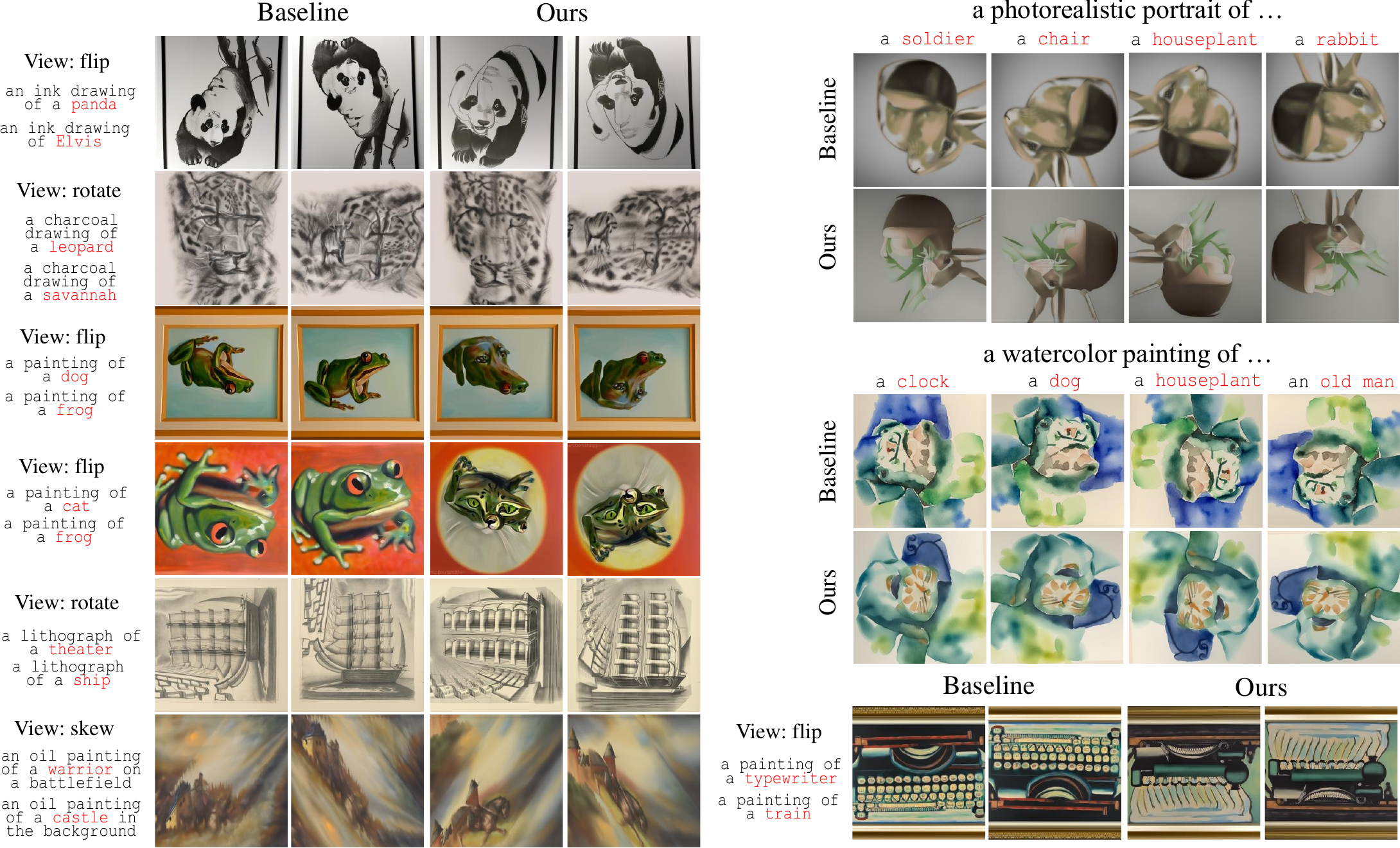}
    \caption{More qualitative results of our proposed method compared to the baseline method\cite{geng2024visual}}
    \vspace{-1mm}
    \label{fig:quali-cmp}
    \vspace{-4mm}
\end{figure*}

\subsection{Anti-Segregation Optimization}
\label{sec:anti-se}

In multi-task learning (MTL), a single model is trained to solve multiple tasks simultaneously, where shared representations are learned to exploit commonalities between tasks\cite{10.1093/nsr/nwx105}. Similarly, in visual anagram generation, concepts from different prompts are expected to share a single image.

However, the lack of explicit control over region sharing in the image can lead to concept segregation, as discussed in \cref{sec:intro}. In such cases, concepts from different prompts may appear as distinct, separate objects (as shown in \cref{fig:fail1}), which is considered a failure in anagram generation.
We conjecture that this is caused by the attention maps of different views being too disjointed. Therefore, we design an inference-time loss term to encourage the intersection of the attention maps of concepts from different views. Specifically, we assume that there is only one concept $c_i$ in each view $i$, which is the main object in the corresponding prompt (e.g., the words in red in \cref{fig:method_overview,fig:quali-cmp}), and the loss term is defined as:
\begin{equation}
\resizebox{0.9\linewidth}{!}{
\begin{math}
\begin{gathered}
\mathcal{L}_{\text{Anti-Seg}} = \frac{1}{N(N-1)}\sum_{1\le i < j \le N}{\left | \phi - \frac{\sum_{p}{\min(A_{t,p}^{c_i}, A_{t,p}^{c_j})}}{\sum_{p}{(A_{t,p}^{c_i}+ A_{t,p}^{c_j})}}\right |},
\label{eq:anti-se}
\end{gathered}
\end{math}
}
\end{equation}
where $p$ iterates over all pixels in the attention maps, $A_{t,p}^{c_i}$ denotes the summed attention score of all tokens derived from concept $c_i$ at pixel $p$ at time step $t$, and $N$ is the number of views.
$\phi$ is a hyperparameter that controls the target overlap ratio between attention maps, and will be discussed in \cref{sec:target}.
At each denoising step, after the denoised image is computed, this loss term is used to modulate it through single-step gradient descent with one step of look-ahead:
\begin{equation}
    x_{t-1}=x_{t-1}' - \alpha\nabla_{x_{t-1}'}\mathcal{L}_{\!\text{Anti-Seg}}
\end{equation}
where $\alpha$ controls the step size. The updated image will be passed to the next denoising step.

\subsection{Noise Vector Balancing}
\label{sec:noise-balancing}

A common challenge in multi-task learning is the imbalance of gradients across tasks, where certain tasks may dominate the learning process \cite{yugradientsurgerymultitask_2020}. To address this, GradNorm \cite{chengradnormgradientnormalization_2018} propose to measure each task's training progress using the ratio of loss decrease, and assigns higher weights to gradients from tasks that have shown less progress.

Building on this idea, we propose reweighting the estimated noise vectors at each time step based on task completion scores. Specifically, text prompts and their viewed corresponding noisy images are respectively fed into a CLIP text encoder and a noise-aware CLIP image encoder, which has been pre-trained on both clean and noisy images. The cosine similarity between the text and image embeddings serves as the task completion score, which is then used to reweight the noise vectors before combining them. The reweighting and combining process can be formally expressed as:\vspace{-1mm}
\begin{align}
    \text{CS}_{t}^{i} &= \cos\left <\mathcal{E_{\text{T}}}(y_i),\mathcal{E_{\text{I}}}(v_i(x_t))\right >\label{eq:comp-scr}\\
    \hat{\alpha_{t}^{i}} &=(\text{CS}_{t}^{i})^{-2+\frac{t}{T}}\label{eq:rew}\\
    \alpha_{t}^{i} &=\frac{\hat{\alpha_{t}^{i}}}{\sum_{j=1}^{N}\hat{\alpha_{t}^{j}}}\label{eq:w-norm}\\
    \tilde{\epsilon_{t}} &= \sum_{i=1}^{N}\alpha_{t}^{i}v_i^{-1}(\epsilon_{t}^{i})\label{eq:comb-noi}
\end{align}
where $\mathcal{E_{\text{T}}}$ and $\mathcal{E_{\text{I}}}$ represent the CLIP text encoder and the noise-aware CLIP image encoder respectively, $i$ indexes the views, $v_i$ denotes the view transformation function, $y_i$ is the text prompt for view $i$, $\epsilon_{t}^{i}$ denotes the noise prediction for view $i$ at time step $t$, and $T$ is the total number of denoising steps. The reweighted noise vector $\tilde{\epsilon_{t}}$ is then used to compute the next image.

Intuitively, in \cref{eq:comp-scr}, a higher cosine similarity indicates better alignment between the text prompt and the image under a given view, suggesting that the generation task is closer to completion.
Therefore, the negative exponent in \cref{eq:rew} assigns lower weights to the noise vectors for that view, preventing it from dominating the denoising process. Additionally, the linear term in the exponent is designed to gradually increase the importance of the task completion score as the denoising process progresses, which is based on the observation that CLIP's task completion score becomes more reliable as the image becomes cleaner. \cref{eq:w-norm} normalizes the reweighted coefficients and \cref{eq:comb-noi} combines the reweighted noise vectors together.

\begin{figure}
    \centering
    \includegraphics[width=0.95\linewidth]{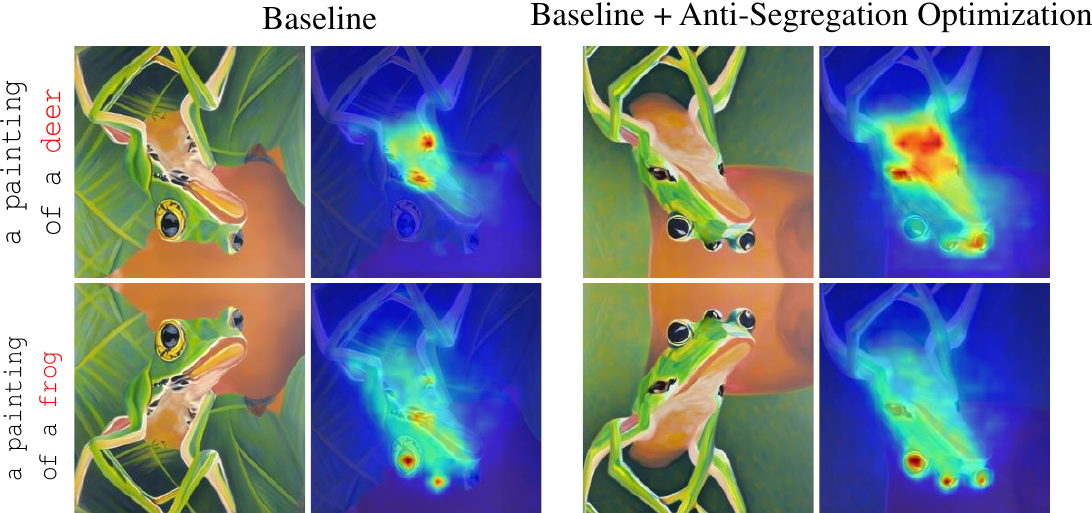}
    \vspace{-1mm}
    \caption{Qualitative results of Anti-Segregation Optimization and visualization of attention maps.}
    \label{fig:deer_and_frog}
    \vspace{-1mm}
\end{figure}

\begin{figure}
    \centering
    \includegraphics[width=1\linewidth]{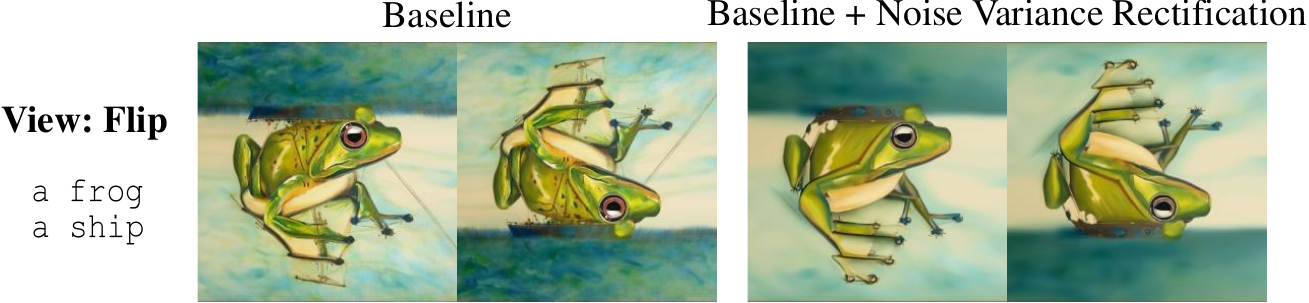}
    \vspace{-5mm}
    \caption{Qualitative results of Noise Variance Rectification.}
    \label{fig:frog_ship}
    \vspace{-3mm}
\end{figure}

\subsection{Noise Variance Rectification}
\label{sec:noise-unitization}

As stated in \cite{hodenoisingdiffusionprobabilistic_2020}, one of the most important assumptions of denoising diffusion models is that the estimated noise vectors follow a Gaussian distribution with zero mean and unit variance, which is crucial for the model to reverse the diffusion process. Existing work \cite{wangpretrainingallyou_2022} has identified that the introduction of classifier-free guidance incurs a statistics shift, which might hinder the denoising process, and proposed normalized classifier-free guidance to rectify this issue. 

In the context of diffusion-based visual anagram generation, although the estimated noise vectors for each view are expected to possess standard Gaussian properties because the model that generates them is trained under this assumption, the combined noise vector may not, which could be detrimental to the denoising process. Therefore, we propose to rectify the combined noise vector to preserve the desired statistical properties. Below we provide a detailed derivation of the rectification process. For convenience, we abuse the notation and denote $v_i^{-1}(\epsilon_{t}^{i})$ as $\epsilon_{t}^{i}$ in this subsection. Basic assumptions include:
\begin{equation}
    \forall i, \epsilon_{t}^{i} \sim \mathcal{N}(0,I)
\end{equation}

Therefore, for every element $p$ in the combined noise vector given by \cref{eq:comb-noi}, we have:
\begin{equation}
    \mathbb{E}[\tilde{\epsilon_{t,p}}] = \mathbb{E}\left [\sum_{i=1}^{N}{\alpha_{t}^{i}\epsilon_{t,p}^{i}}\right ] = \sum_{i=1}^{N}{\alpha_{t}^{i}\mathbb{E}[\epsilon_{t,p}^{i} ]} = 0
\end{equation}

\begin{figure}
    \centering
    \includegraphics[width=0.95\linewidth]{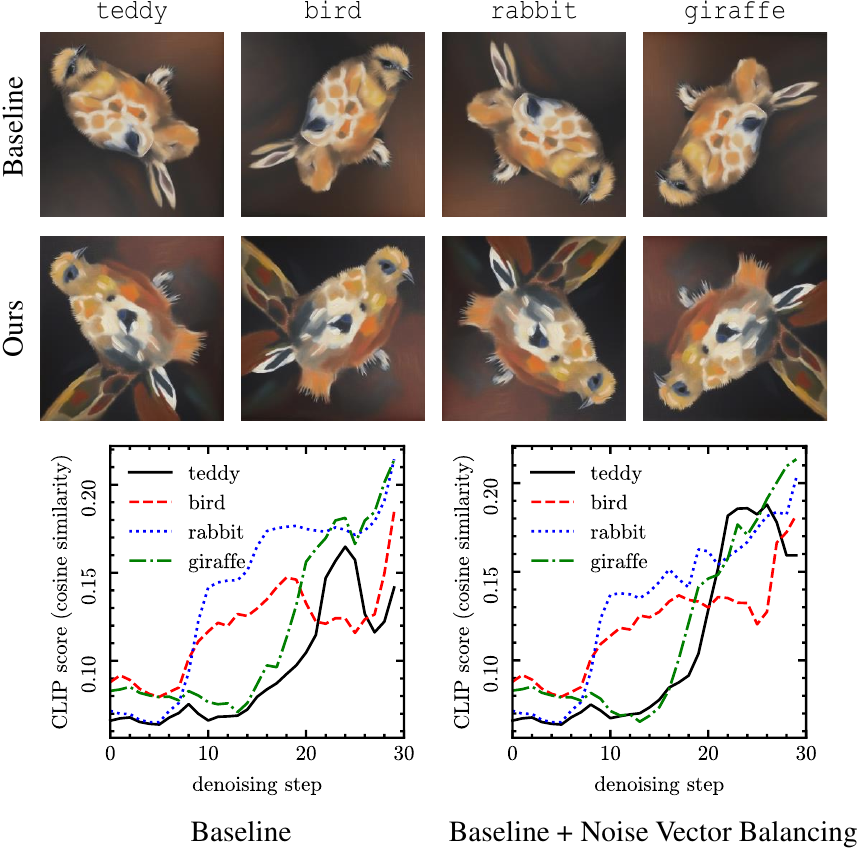}
    \vspace{-2mm}
    \caption{Qualitative results of Noise Vector Balancing and visualization of the task completion curve.}
    \label{fig:teddy_rabbit_bird_giraffe}
    \vspace{-4mm}
\end{figure}

This suggests that every element in the combined noise vector still has an expectation of zero. However, the variance is not guaranteed to be one. To address this issue, we propose to rectify the combined noise vector by applying a scale factor $c$ to make the variance unit. Essentially, we hope to find a scale factor $c$ that satisfies:\vspace{-4mm}
\begin{equation}
\resizebox{.9\linewidth}{!}{
\begin{math}
\begin{aligned}
        &\mathbb{D}[c\cdot\tilde{\epsilon_{t,p}}] = c^2\mathbb{D}[\tilde{\epsilon_{t,p}}] =c^2\mathbb{D}\left [\sum_{i=1}^{N}{\alpha_{t}^{i}\epsilon_{t,p}^{i}}\right ]\\
        =&c^2\left [ \sum_{i=1}^{N}{(\alpha_{t}^{i})^2\mathbb{D}[\epsilon_{t,p}^{i}]} +\sum_{1\le i<j\le N}{2\alpha_{t}^{i}\alpha_{t}^{j}\text{Cov}(\epsilon_{t,p}^{i},\epsilon_{t,p}^{j})}\right ] =1
\end{aligned}
\end{math}
}
\end{equation}
since we have $\mathbb{D}[\epsilon_{t,p}^{i}]=1$, the scale factor $c$ can be computed as:\vspace{-3mm}
\begin{equation}
\resizebox{.85\linewidth}{!}{
\begin{math}
\begin{gathered}
    c=\frac{1}{\sqrt{\sum_{i=1}^{N}{(\alpha_{t}^{i})^2}+\sum_{1\le i<j\le N}{2\alpha_{t}^{i}\alpha_{t}^{j}\text{Cov}(\epsilon_{t,p}^{i},\epsilon_{t,p}^{j})}}}
    \label{eq:scale}
\end{gathered}
\end{math}
}
\end{equation}
where the covariance term satisfies:
\begin{equation}
\resizebox{.85\linewidth}{!}{
\begin{math}
\begin{gathered}
    \text{Cov}(\epsilon_{t,p}^{i},\epsilon_{t,p}^{j}) = \mathbb{E}[\epsilon_{t,p}^{i}\epsilon_{t,p}^{j}] - \mathbb{E}[\epsilon_{t,p}^{i}]\mathbb{E}[\epsilon_{t,p}^{j}] = \mathbb{E}[\epsilon_{t,p}^{i}\epsilon_{t,p}^{j}]
\end{gathered}
\end{math}
}
\end{equation}
since both $\epsilon_{t,p}^{i}$ and $\epsilon_{t,p}^{j}$ follow standard Gaussian, the covariance term equals their correlation coefficient. We assume that within the same denoising step, the correlation coefficients between different views' noise vectors are the same across all elements. Let $C\times H \times W$ be the total number of elements in each noise vector, which is usually large, so according to the law of large numbers, we can approximate the covariance term as:
\begin{equation}
    \rho_{i,j} = \text{Cov}(\epsilon_{t,p}^{i},\epsilon_{t,p}^{j}) = \mathbb{E}[\epsilon_{t,p}^{i}\epsilon_{t,p}^{j}] \approx \frac{\sum_{p}{\epsilon_{t,p}^{i}\epsilon_{t,p}^{j}}}{C\times H \times W}
    \label{eq:cov}
\end{equation}

Now we can compute the desired scale factor $c$ by plugging in the estimated covariance term. The rectified noise vector is just the combined noise vector multiplied by this scale factor.

\section{Experiments}

\subsection{Implementation Details}

We utilize DeepFloyd \cite{stabilityaireleases_} as our backbone diffusion model, a cutting-edge open-source text-to-image diffusion model.
It operates in the pixel space instead of latent space, which is crucial for producing high-quality visual anagrams as \cite{gengvisualanagramsgenerating_2023} mentioned.
Our evaluations focus on images generated at the second stage of DeepFloyd, with a resolution of 256 $\times$ 256. 
For task completion measurement, we employ the ViT-B CLIP mode released by \cite{nicholglidephotorealisticimage_2022}, which is pre-trained on both noisy and clean images.
Our proposed Anti-Segregation Optimization is applied specifically to the initial stage of DeepFloyd, which is responsible for generating 64 $\times$ 64 images from pure noise and determining the general layout of the image, where concept segregation issues often arise.
Noise vector balancing and variance rectification are implemented in both image generation and super-resolution stages.
It is important to note that training or fine-tuning of the models is not required, and indeed, a single RTX 3090 suffices to run our proposed pipeline.

\subsection{Dataset and Metrics}

For quantitative evaluation and ablation study, we compile a 2-view dataset with text prompt pairs using 10 classes of objects from the CIFAR-10 dataset \cite{Krizhevsky2009LearningML}, that generally follows the setup in \cite{gengvisualanagramsgenerating_2023}. We further extend the dataset to 3-view to evaluate the scalability of our proposed method. For each pair (or triplet) of text prompts, we generate 10 images with different seeds and report the average results.

\begin{table}[t]
    \centering
    \footnotesize
    \resizebox{.9\columnwidth}{!}{
        \begin{tabular}{c|l|ccc}
            \toprule
            Dataset                     & Method                                            & $\mathcal{A}_{\!\text{min}}\uparrow$ & $\mathcal{C}\uparrow$ & $\mathcal{A}_{\!\text{avg}}\uparrow$ \\
            \midrule
             & Tancik\cite{tancik_tancikillusion-diffusion_2024} & 0.1997                               & 0.5568                & 0.2295                               \\
                                   CIFAR10     & Burgert \etal \cite{burgert_diffusion_2024}       & 0.2547                               & 0.6867                & 0.2676                               \\
                                   2-views     & Visual Anagrams\cite{geng2024visual}                     & 0.2583                               & 0.6744                & 0.2717                               \\
                                        & Ours                                              & \textbf{0.2711}                      & \textbf{0.6913}       & \textbf{0.2816}                      \\
            \midrule
             & Tancik\cite{tancik_tancikillusion-diffusion_2024} & 0.1914                               & 0.4043                & 0.2207                               \\
                            CIFAR10        & Burgert \etal \cite{burgert_diffusion_2024}       & N/A                                    & N/A                     & N/A                                    \\
                               3-view   & Visual Anagrams \cite{geng2024visual}                     & 0.2363                               & 0.4892                & 0.2642                               \\
                                        & Ours                                              & \textbf{0.2507}                      & \textbf{0.4955}       & \textbf{0.2759}                      \\
            \bottomrule
        \end{tabular}
    }
    \vspace{-1mm}
    \caption{\textbf{Quantitative Results.} We report the results of our proposed method against baseline\cite{geng2024visual}, together with other existing methods on the 2-view dataset and the 3-view dataset. Burgert \etal \cite{burgert_diffusion_2024} only supports 2 views.}
    \vspace{-2mm}
    \label{tab:quati}
\end{table}

\begin{table}[t]
    \centering
    \footnotesize
    \resizebox{0.95\linewidth}{!}{
        \begin{tabular}{c|ccc|ccc}
            \toprule
            Method   & ASO        & NVB        & NVR        & $\mathcal{A}_{\min}\uparrow$ & $\mathcal{C}\uparrow$ & $\mathcal{A}_{avg}\uparrow$ \\
            \midrule
            Baseline & $\times$   & $\times$   & $\times$   & 0.2583                       & 0.6744                & 0.2717                      \\
            \midrule
                     & \checkmark & $\times$   & $\times$   & 0.2594                       & 0.6821                & 0.2716                      \\
                     & $\times$   & \checkmark & $\times$   & 0.2632                       & 0.6864                & 0.2734                      \\
                     & $\times$   & $\times$   & \checkmark & 0.2642                       & 0.6739                & 0.2784                      \\
                     & $\times$   & \checkmark & \checkmark & 0.2697                       & 0.6834                & 0.2807                      \\
                     & \checkmark & $\times$   & \checkmark & 0.2662                       & 0.6816                & 0.2795                      \\
                     & \checkmark & \checkmark & $\times$   & 0.2628                       & 0.6913                & 0.2734                      \\
            \midrule
            Full     & \checkmark & \checkmark & \checkmark & \textbf{0.2711}              & \textbf{0.6913}       & \textbf{0.2816}             \\
            \bottomrule
        \end{tabular}
    }
    \vspace{-1mm}
    \caption{\textbf{Ablation Study.} We ablate our proposed Anti-Segregation Optimization (ASO), Noise Vector Balancing (NVB), and Noise Variance Rectification (NVR) techniques.}
    \vspace{-3mm}
    \label{tab:ablation}
\end{table}

We follow \cite{gengvisualanagramsgenerating_2023} and compute a score matrix $S\in \mathcal{R}^{N\times N}$ for each generated image, where $N$ is the number of views, and $S_{i,j}$ is the cosine similarity between the CLIP text embedding of the $i$-th text prompt and the CLIP image embedding of the $j$-th view of the generated image. We then compute three metrics to evaluate the quality of the generated visual anagrams: 

\begin{enumerate}
    \vspace{-2mm}
    \item \textbf{Worst Alignment Score ($\mathcal{A}_{\!\text{min}}$)}: The minimum value in the diagonal of the score matrix $S$, which measures the worst alignment between the text prompts and the generated image across all views.\vspace{-2mm}
    \item \textbf{Concealment Score ($\mathcal{C}$)}: $\mathcal{C} = \frac{1}{N}\text{tr}(\!\text{softmax}(\frac{S}{\tau}))$, where $\text{tr}(\cdot)$ denotes the trace of a matrix, $\tau$ is the temperature parameter of CLIP, and $\text{softmax}$ is computed against both rows and columns of $S$. This metric measures how well the generated image conceals concepts in other text prompts under the current view.\vspace{-2mm}
    \item \textbf{Average Alignment Score ($\mathcal{A}_{\!\text{avg}}$)}: The average value in the diagonal of the score matrix $S$, which measures the overall alignment across all views.\vspace{-2mm}
\end{enumerate}

Actually, $\mathcal{A}_{\!\text{min}}$ and $\mathcal{C}$ are proposed by \cite{geng2024visual}, we further introduce $\mathcal{A}_{\!\text{avg}}$ to demonstrate that our method not only balances the trade-off between different views but also boosts the overall text-image alignment. Note that we use a vanilla CLIP model that is only pre-trained on clean images to compute the metrics, which is different from the one used in our proposed method and is more suitable for evaluating the generated images, and also prevents our method from directly optimizing the metrics.

\subsection{Results}

\begin{figure}
    \centering
    \begin{subfigure}{.49\linewidth}
        \includegraphics[width=\linewidth]{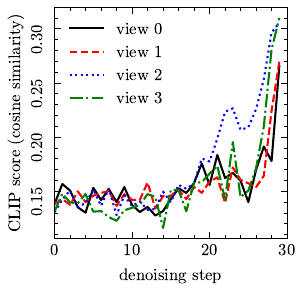}
        \caption{Vanilla CLIP}
        \vspace{-1mm}
    \end{subfigure}
        \begin{subfigure}{.49\linewidth}
        \includegraphics[width=\linewidth]{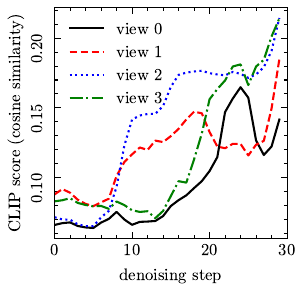}
        \caption{Noise-aware CLIP}
        \vspace{-1mm}
    \end{subfigure}
    \caption{\textbf{Task completion scores comparison.} We compare the task completion scores measured by noise-aware and vanilla CLIP models during the denoising process using the baseline method.}
    \vspace{-1mm}
    \label{fig:clip_sel}
\end{figure}

\begin{table}[t]
    \centering
    \footnotesize
    \resizebox{\linewidth}{!}{
        \begin{tabular}{ccc|ccc}
            \toprule
            \multicolumn{1}{l}{NVB with}         & \multicolumn{1}{l}{NVB  with}    & \multicolumn{1}{l}{ASO}    & \multirow{2}{*}{$\mathcal{A}_{\min}\uparrow$} & \multirow{2}{*}{$\mathcal{C}\uparrow$} & \multirow{2}{*}{$\mathcal{A}_{avg}\uparrow$} \\
            \multicolumn{1}{l}{Noise-aware CLIP} & \multicolumn{1}{l}{Vanilla CLIP} & \multicolumn{1}{l}{\& NVR} &                                               &                                        &                                              \\
            \midrule
            $\times$                             & \checkmark                       & $\times$                   & 0.2609                                        & \textbf{0.6868}                        & 0.2726                                       \\
            \checkmark                           & $\times$                         & $\times$                   & \textbf{0.2632}                               & 0.6864                                 & \textbf{0.2734}                              \\
            \midrule
            $\times$                             & \checkmark                       & \checkmark                 & 0.2690                                        & 0.6838                                 & 0.2808                                       \\
            \checkmark                           & $\times$                         & \checkmark                 & \textbf{0.2711}                               & \textbf{0.6913}                        & \textbf{0.2816}                              \\
            \bottomrule
        \end{tabular}
    }
    \caption{\textbf{Ablation Study.} We compare the performance of our proposed method using different CLIP models to measure task completion.}
    \vspace{-3mm}
    \label{tab:clip-sel}
\end{table}

\textbf{Quantitative Results.}
We first report quantitative results of our proposed method against the baseline method, together with other existing methods on the 2-view dataset and the 3-view dataset in \cref{tab:quati}. For the 2-view dataset, identity and vertical flip are used as the two views, as they are supported by all methods. For the 3-view dataset, we add clockwise rotation as the third view. A total of 450 images are generated by each method for the 2-view dataset, and 1200 images for the 3-view dataset. We compare our method with the baseline method that straightforwardly averages the noise predictions from different text prompts, and the results demonstrate that our method outperforms the baseline method across all metrics on both datasets.

\textbf{Qualitative Results.} We also show qualitative results of our proposed methods to demonstrate their effectiveness.
\cref{fig:deer_and_frog} shows an example when we use ``a painting of a deer'' and ``a painting of a frog'' as the text prompts to generate visual anagrams under the identity view and the vertical flip view. In the image generated by the baseline method, we could only see a small deer head in the identity view, where attention activates, and the rest of the image does not make sense for the deer prompt. In contrast, after applying our proposed Anti-Segregation Optimization, the denoising process learns to find commonalities between the two generation tasks and generates antlers in the identity view that can also be interpreted as frog legs in the flipped view, generates a deer head that can also be interpreted as the frog's body, and now the upper-right background region in the flipped view naturally interprets as the part of the deer's body in the identity view. 

In \cref{fig:teddy_rabbit_bird_giraffe}, we provide an example to compare the generation process and results before and after applying our proposed Noise Vector Balancing technique. We use ``an oil painting of'' as the style prefix and ``a teddy'', ``a bird'', ``a rabbit'', and ``a giraffe'' as the text prompts to generate 4-view visual anagrams. As can be seen, in the image generated by the baseline method, the teddy bear is hardly recognizable in the identity view because the generation process is dominated by other tasks. After applying our proposed method, the teddy bear is more recognizable, and the task completion score curve also shows that the gap between the teddy and the other tasks is reduced, which demonstrates that our method effectively balances multiple generation tasks.
\cref{fig:frog_ship} illustrates the necessity to preserve the variance of noise vectors during the denoising process. After applying Noise Variance Rectification, artifacts and distortions are reduced, and the overall quality of the generated image is improved. \cref{fig:quali-cmp} shows more qualitative results, where our method generates more visually appealing images with fewer artifacts, and better align with text prompts, compared to the baseline method.

\begin{figure}
    \centering
    \includegraphics[width=\linewidth]{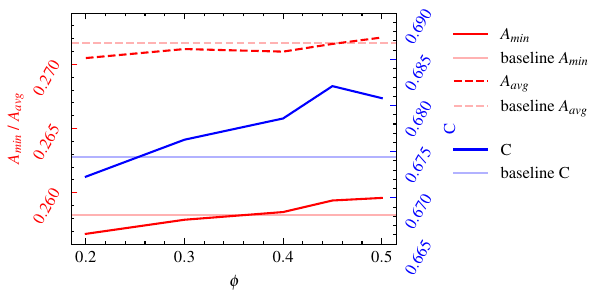}
    \vspace{-8mm}
    \caption{Effect of target overlap ratio in Anti-Segregation Optimization.}
    \vspace{-3mm}
    \label{fig:phi}
\end{figure}

\begin{figure}
    \centering
    \begin{subfigure}{.49\linewidth}
        \includegraphics[width=\linewidth]{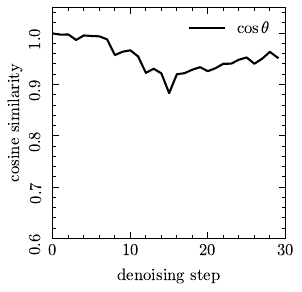}
        \caption{Cosine Similarity}
        \vspace{-2mm}
    \end{subfigure}
        \begin{subfigure}{.49\linewidth}
        \includegraphics[width=\linewidth]{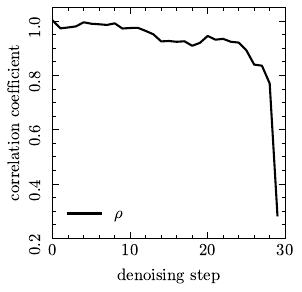}
        \caption{Correlation Coefficient}
        \vspace{-2mm}
    \end{subfigure}
    \caption{\textbf{Correlation between noise vectors.} We record the cosine similarity and estimated correlation coefficient between the predicted noise vectors of different views at each denoising step.}
    \vspace{-3mm}
    \label{fig:noi_cor}
\end{figure}

\subsection{Ablation Study}

We ablate our proposed Anti-Segregation Optimization and Noise Vector Balancing techniques, together with the Noise Variance Rectification in \cref{tab:ablation}. The results show that each component contributes to the overall performance of our method, and the combination of all components achieves the best results. Specifically, the Anti-Segregation Optimization effectively prevents concept segregation and primarily contributes to the improvement of $\mathcal{C}$, the Noise Vector Balancing technique balances the trade-off between different views and thus improves $\mathcal{A}_{\text{min}}$ and $\mathcal{C}$, and the Noise Variance Rectification corrects the denoising process and boosts overall text-image alignment.

\section{Additional Results and Discussion}
\vspace{-1mm}

\subsection{Comparison with Vanilla CLIP Models}
\label{sec:clip-sel}
Existing works on leveraging auxiliary models for text-to-image generation, such as classifier guidance and CLIP guidance, have been shown to work better when the auxiliary models are pre-trained on noisy images,
otherwise the intermediate images are out-of-distribution for the auxiliary models.
Therefore, in our proposed method, we use a noise-aware CLIP model to measure the degree of task completion in the noise vector balancing process. Here we provide further analysis on this design choice.

\cref{fig:clip_sel} depicts a typical example of task completion scores measured by noise-aware and vanilla CLIP models during the denoising process when we use the baseline method to generate a 4-view visual anagram. We can observe that curves of task completion scores measured by the noise-aware CLIP model are smoother and less noisy compared to those measured by a vanilla CLIP model, which indicates that the noise-aware CLIP model provides more stable and more reliable task completion scores. 
Additionally, the noise-aware CLIP model starts to provide useful information earlier than the vanilla CLIP model, which helps the noise vector balancing process to take early actions to prevent concept domination. Furthermore, we also compare the performance of our proposed method using noise-aware and vanilla CLIP models in \cref{tab:clip-sel}, and the results show that our method using noise-aware CLIP model generally outperforms the one using vanilla CLIP model, which further validates our design choice.

\subsection{Effect of Target Overlap Ratio in Anti-Segregation Optimization}
\label{sec:target}
In \cref{eq:anti-se}, we introduce a hyperparameter $\phi$, which describes the desired overlap ratio between attention maps of different views and takes values in the range $[0, 0.5]$. Generally, a larger $\phi$ encourages more overlap between attention maps, while a smaller $\phi$ encourages them to be more segregated, and $\phi=0.5$ favors complete overlap.
We investigate the effect of $\phi$ on the performance of our proposed Anti-Segregation Optimization (ASO) with other techniques disabled on the 2-view dataset and the results are shown in \cref{fig:phi}. We can observe that our method achieves the best performance around $\phi=0.45$.

\vspace{-1mm}
\subsection{Correlation between Noise Vectors}
\vspace{-1mm}

While it is true that statistical property of the noise vectors is crucial for the denoising diffusion models to reverse the diffusion process, one may question why simply averaging the noise predictions from different text prompts still has a chance to generate sensible images, because averaging $N$ independent standard Gaussian noise vectors should result in a noise vector with zero mean and variance of $1/N$, and such significant variance shift seems to be fatal.

Therefore, we investigate the correlation between the noise vectors at each denoising step by recording cosine similarities and estimated correlation coefficients using \cref{eq:cov} between them.
\cref{fig:noi_cor} shows a typical example of the denoising process, when we use the baseline method to generate a 2-view visual anagram, and we can observe that the correlation coefficient and cosine similarity are close to one. Therefore, in \cref{eq:scale}, where $\alpha_t^i=\frac{1}{N}$ for the baseline method, the scale factor $c$ is close to one, which explains why the baseline method is still possible to generate images that do not drift too far from the data distribution. Anyhow, our proposed method is beneficial for better generating visual anagrams, as it explicitly corrects the denoising process.

\vspace{-2mm}
\section{Conclusions and Limitations}
\vspace{-2mm}
In this work, we link visual anagram generation with multi-task learning and draw inspiration from the latter to address the challenges of concept separation and concept domination in the former. We propose an Anti-Segregation Optimization technique that encourages concepts in different views to overlap, and a Noise Vector Balancing technique that reweights the noise predictions based on task completion scores to balance the denoising process. We further propose to scale the combined noise vector to preserve the variance of the noise. Our experiments demonstrate that our proposed method outperforms the baseline method consistently across different metrics and datasets.

Despite the promising results, our work has several limitations. Firstly, we are still unable to leverage powerful latent space diffusion models like Stable Diffusion to generate good visual anagrams because latents and the decoded images are not consistent across transformations as discussed in \cite{geng2024visual}. Secondly, our method could not extend anagram generation to non-orthogonal transformations like non-volume-preserving deformations, which is an interesting direction for future work.

{\small
\bibliographystyle{ieee_fullname}
\bibliography{main}

\begin{thebibliography}{10}\itemsep=-1pt

\bibitem{stabilityaireleases_}
Stability ai releases deepfloyd if, a powerful text-to-image model that can
  smartly integrate text into images.
\newblock https://stability.ai/news/deepfloyd-if-text-to-image-model.

\bibitem{agarwalastartesttimeattention_2023}
Aishwarya Agarwal, Srikrishna Karanam, K.~J. Joseph, Apoorv Saxena, Koustava
  Goswami, and Balaji~Vasan Srinivasan.
\newblock A-star: Test-time attention segregation and retention for
  text-to-image synthesis, 2023.

\bibitem{baoseparateenhancecompositionalfinetuning_2024}
Zhipeng Bao, Yijun Li, Krishna~Kumar Singh, Yu-Xiong Wang, and Martial Hebert.
\newblock Separate-and-enhance: Compositional finetuning for text2image
  diffusion models, 2024.

\bibitem{boring1930new}
Edwin~G Boring.
\newblock A new ambiguous figure.
\newblock {\em The American Journal of Psychology}, 1930.

\bibitem{burgert_diffusion_2024}
Ryan Burgert, Xiang Li, Abe Leite, Kanchana Ranasinghe, and Michael Ryoo.
\newblock Diffusion {Illusions}: {Hiding} {Images} in {Plain} {Sight}.
\newblock In {\em {ACM} {SIGGRAPH} 2024 {Conference} {Papers}}, {SIGGRAPH} '24,
  pages 1--11, New York, NY, USA, July 2024. Association for Computing
  Machinery.

\bibitem{chandradesigningperceptualpuzzles_2022}
Kartik Chandra, Tzu-Mao Li, Joshua Tenenbaum, and Jonathan {Ragan-Kelley}.
\newblock Designing perceptual puzzles by differentiating probabilistic
  programs.
\newblock In {\em Special Interest Group on Computer Graphics and Interactive
  Techniques Conference Proceedings}, pages 1--9, 2022.

\bibitem{cheferattendexciteattentionbasedsemantic_2023}
Hila Chefer, Yuval Alaluf, Yael Vinker, Lior Wolf, and Daniel {Cohen-Or}.
\newblock Attend-and-excite: Attention-based semantic guidance for
  text-to-image diffusion models, 2023.

\bibitem{chen2024ultraman}
Mingjin Chen, Junhao Chen, Xiaojun Ye, Huan-ang Gao, Xiaoxue Chen, Zhaoxin Fan,
  and Hao Zhao.
\newblock Ultraman: Single image 3d human reconstruction with ultra speed and
  detail.
\newblock {\em arXiv preprint arXiv:2403.12028}, 2024.

\bibitem{chen2022cerberus}
Xiaoxue Chen, Tianyu Liu, Hao Zhao, Guyue Zhou, and Ya-Qin Zhang.
\newblock Cerberus transformer: Joint semantic, affordance and attribute
  parsing.
\newblock In {\em Proceedings of the IEEE/CVF Conference on Computer Vision and
  Pattern Recognition}, pages 19649--19658, 2022.

\bibitem{9682598}
Xiaoxue Chen, Hao Zhao, Guyue Zhou, and Ya-Qin Zhang.
\newblock Pq-transformer: Jointly parsing 3d objects and layouts from point
  clouds.
\newblock {\em IEEE Robotics and Automation Letters}, 7(2):2519--2526, 2022.

\bibitem{chengradnormgradientnormalization_2018}
Zhao Chen, Vijay Badrinarayanan, Chen-Yu Lee, and Andrew Rabinovich.
\newblock Gradnorm: Gradient normalization for adaptive loss balancing in deep
  multitask networks, 2018.

\bibitem{chenjustpicksign_2020}
Zhao Chen, Jiquan Ngiam, Yanping Huang, Thang Luong, Henrik Kretzschmar, Yuning
  Chai, and Dragomir Anguelov.
\newblock Just pick a sign: Optimizing deep multitask models with gradient sign
  dropout.
\newblock https://arxiv.org/abs/2010.06808v1, 2020.

\bibitem{chucamouflageimages_2010}
Hung~Kuo Chu, Wei~Hsin Hsu, Niloy~J. Mitra, Daniel {Cohen-Or}, Tien-Tsin Wong,
  and Tong-Yee Lee.
\newblock Camouflage images.
\newblock {\em ACM Transactions on Graphics}, 29(4):51, 2010.

\bibitem{dhariwaldiffusionmodelsbeat_2021}
Prafulla Dhariwal and Alex Nichol.
\newblock Diffusion models beat gans on image synthesis.
\newblock https://arxiv.org/abs/2105.05233v4, 2021.

\bibitem{ferino2017arcimboldo}
Sylvia Ferino-Pagden.
\newblock {\em Arcimboldo}.
\newblock Skira, 2017.

\bibitem{freemanmotionmovement_1991}
William~T. Freeman, Edward~H. Adelson, and David~J. Heeger.
\newblock Motion without movement.
\newblock {\em ACM SIGGRAPH Computer Graphics}, 25(4):27--30, 1991.

\bibitem{gao2024scp}
Huan-ang Gao, Mingju Gao, Jiaju Li, Wenyi Li, Rong Zhi, Hao Tang, and Hao Zhao.
\newblock Scp-diff: Photo-realistic semantic image synthesis with
  spatial-categorical joint prior.
\newblock {\em arXiv preprint arXiv:2403.09638}, 2024.

\bibitem{gengvisualanagramsgenerating_2023}
Daniel Geng, Inbum Park, and Andrew Owens.
\newblock Visual anagrams: Generating multi-view optical illusions with
  diffusion models.
\newblock https://arxiv.org/abs/2311.17919v2, 2023.

\bibitem{geng2024visual}
Daniel Geng, Inbum Park, and Andrew Owens.
\newblock Visual anagrams: Generating multi-view optical illusions with
  diffusion models.
\newblock In {\em Proceedings of the IEEE/CVF Conference on Computer Vision and
  Pattern Recognition}, pages 24154--24163, 2024.

\bibitem{gomez-villaconvolutionalneuralnetworks_2019}
Alexander {Gomez-Villa}, Adrian Martin, Javier {Vazquez-Corral}, and Marcelo
  Bertalmio.
\newblock Convolutional neural networks can be deceived by visual illusions.
\newblock In {\em 2019 IEEE/CVF Conference on Computer Vision and Pattern
  Recognition (CVPR)}, pages 12301--12309, Long Beach, CA, USA, 2019.

\bibitem{graikosdiffusionmodelsplugplay_2023}
Alexandros Graikos, Nikolay Malkin, Nebojsa Jojic, and Dimitris Samaras.
\newblock Diffusion models as plug-and-play priors, 2023.

\bibitem{guizilinisemanticallyguidedrepresentationlearning_2020}
Vitor Guizilini, Rui Hou, Jie Li, Rares Ambrus, and Adrien Gaidon.
\newblock Semantically-guided representation learning for self-supervised
  monocular depth.
\newblock https://arxiv.org/abs/2002.12319v1, 2020.

\bibitem{guodynamictaskprioritization_2018a}
Michelle Guo, Albert Haque, De-An Huang, Serena Yeung, and Li {Fei-Fei}.
\newblock Dynamic task prioritization for multitask learning.
\newblock In Vittorio Ferrari, Martial Hebert, Cristian Sminchisescu, and Yair
  Weiss, editors, {\em Computer Vision -- ECCV 2018}, pages 282--299, Cham,
  2018.

\bibitem{guoganmouflage3dobject_2023}
Rui Guo, Jasmine Collins, Oscar {de Lima}, and Andrew Owens.
\newblock Ganmouflage: 3d object nondetection with texture fields, 2023.

\bibitem{hertzmannvisualindeterminacygan_2020}
Aaron Hertzmann.
\newblock Visual indeterminacy in gan art.
\newblock {\em Leonardo}, 53(4):424--428, 2020.

\bibitem{hodenoisingdiffusionprobabilistic_2020}
Jonathan Ho, Ajay Jain, and Pieter Abbeel.
\newblock Denoising diffusion probabilistic models, 2020.

\bibitem{igarashimultitaskcurriculumlearning_2022}
Hiroaki Igarashi.
\newblock Multi-task curriculum learning based on gradient similarity.
\newblock {\em BMVC}, 2022.

\bibitem{jainiintriguingpropertiesgenerative_2024}
Priyank Jaini, Kevin Clark, and Robert Geirhos.
\newblock Intriguing properties of generative classifiers, 2024.

\bibitem{kendallmultitasklearningusing_2018}
Alex Kendall, Yarin Gal, and Roberto Cipolla.
\newblock Multi-task learning using uncertainty to weigh losses for scene
  geometry and semantics, 2018.

\bibitem{Krizhevsky2009LearningML}
Alex Krizhevsky.
\newblock Learning multiple layers of features from tiny images.
\newblock 2009.

\bibitem{li2024fairdiff}
Wenyi Li, Haoran Xu, Guiyu Zhang, Huan-ang Gao, Mingju Gao, Mengyu Wang, and
  Hao Zhao.
\newblock Fairdiff: Fair segmentation with point-image diffusion.
\newblock In {\em International Conference on Medical Image Computing and
  Computer-Assisted Intervention}, pages 617--628. Springer, 2024.

\bibitem{linparetomultitasklearning_2019}
Xi Lin, Hui-Ling Zhen, Zhenhua Li, Qingfu Zhang, and Sam Kwong.
\newblock Pareto multi-task learning, 2019.

\bibitem{liucompositionalvisualgeneration_2022}
Nan Liu, Shuang Li, Yilun Du, Antonio Torralba, and Joshua~B. Tenenbaum.
\newblock Compositional visual generation with composable diffusion models.
\newblock https://arxiv.org/abs/2206.01714v6, 2022.

\bibitem{lu_closed-form_2019}
Ming Lu, Hao Zhao, Anbang Yao, Yurong Chen, Feng Xu, and Li Zhang.
\newblock A {Closed}-{Form} {Solution} to {Universal} {Style} {Transfer}.
\newblock In {\em 2019 {IEEE}/{CVF} {International} {Conference} on {Computer}
  {Vision} ({ICCV})}, pages 5951--5960, Seoul, Korea (South), Oct. 2019. IEEE.

\bibitem{lu2017decoder}
Ming Lu, Hao Zhao, Anbang Yao, Feng Xu, Yurong Chen, and Li Zhang.
\newblock Decoder network over lightweight reconstructed feature for fast
  semantic style transfer.
\newblock In {\em Proceedings of the IEEE international conference on computer
  vision}, pages 2469--2477, 2017.

\bibitem{misracrossstitchnetworksmultitask_2016}
Ishan Misra, Abhinav Shrivastava, Abhinav Gupta, and Martial Hebert.
\newblock Cross-stitch networks for multi-task learning, 2016.

\bibitem{navonmultitasklearningbargaining_2022}
Aviv Navon, Aviv Shamsian, Idan Achituve, Haggai Maron, Kenji Kawaguchi, Gal
  Chechik, and Ethan Fetaya.
\newblock Multi-task learning as a bargaining game.
\newblock https://arxiv.org/abs/2202.01017v2, 2022.

\bibitem{ngoclipfooledoptical_2023a}
Jerry Ngo, S. Sankaranarayanan, and Phillip Isola.
\newblock Is clip fooled by optical illusions?
\newblock In {\em Tiny Papers @ ICLR}, 2023.

\bibitem{nicholglidephotorealisticimage_2022}
Alex Nichol, Prafulla Dhariwal, Aditya Ramesh, Pranav Shyam, Pamela Mishkin,
  Bob McGrew, Ilya Sutskever, and Mark Chen.
\newblock Glide: Towards photorealistic image generation and editing with
  text-guided diffusion models, 2022.

\bibitem{nicholls_perception_2018}
Michael E.~R. Nicholls, Owen Churches, and Tobias Loetscher.
\newblock Perception of an ambiguous figure is affected by own-age social
  biases.
\newblock {\em Scientific Reports}, 8(1):12661, Aug. 2018.
\newblock Publisher: Nature Publishing Group.

\bibitem{hybridimages}
Aude Oliva, Antonio Torralba, and Philippe~G. Schyns.
\newblock Hybrid images.
\newblock {\em ACM Trans. Graph.}, 25(3):527--532, 2006.

\bibitem{owenscamouflagingobjectmany_2014}
Andrew Owens, Connelly Barnes, Alex Flint, Hanumant Singh, and William Freeman.
\newblock Camouflaging an object from many viewpoints.
\newblock In {\em 2014 IEEE Conference on Computer Vision and Pattern
  Recognition}, pages 2782--2789, 2014.

\bibitem{poole2022dreamfusion}
Ben Poole, Ajay Jain, Jonathan~T Barron, and Ben Mildenhall.
\newblock Dreamfusion: Text-to-3d using 2d diffusion.
\newblock {\em arXiv preprint arXiv:2209.14988}, 2022.

\bibitem{rameshzeroshottextimagegeneration_2021}
Aditya Ramesh, Mikhail Pavlov, Gabriel Goh, Scott Gray, Chelsea Voss, Alec
  Radford, Mark Chen, and Ilya Sutskever.
\newblock Zero-shot text-to-image generation.
\newblock https://arxiv.org/abs/2102.12092v2, 2021.

\bibitem{rombachhighresolutionimagesynthesis_2021}
Robin Rombach, Andreas Blattmann, Dominik Lorenz, Patrick Esser, and Bj{\"o}rn
  Ommer.
\newblock High-resolution image synthesis with latent diffusion models.
\newblock https://arxiv.org/abs/2112.10752v2, 2021.

\bibitem{ronneberger2015u}
Olaf Ronneberger, Philipp Fischer, and Thomas Brox.
\newblock U-net: Convolutional networks for biomedical image segmentation.
\newblock In {\em Medical image computing and computer-assisted
  intervention--MICCAI 2015: 18th international conference, Munich, Germany,
  October 5-9, 2015, proceedings, part III 18}, pages 234--241. Springer, 2015.

\bibitem{sahariaphotorealistictextimagediffusion_2022}
Chitwan Saharia, William Chan, Saurabh Saxena, Lala Li, Jay Whang, Emily
  Denton, Seyed Kamyar~Seyed Ghasemipour, Burcu~Karagol Ayan, S.~Sara Mahdavi,
  Rapha~Gontijo Lopes, Tim Salimans, Jonathan Ho, David~J. Fleet, and Mohammad
  Norouzi.
\newblock Photorealistic text-to-image diffusion models with deep language
  understanding, 2022.

\bibitem{senermultitasklearningmultiobjective_2019}
Ozan Sener and Vladlen Koltun.
\newblock Multi-task learning as multi-objective optimization, 2019.

\bibitem{senushkinindependentcomponentalignment_2023}
Dmitry Senushkin, Nikolay Patakin, Arseny Kuznetsov, and Anton Konushin.
\newblock Independent component alignment for multi-task learning, 2023.

\bibitem{sohl-dicksteindeepunsupervisedlearning_2015}
Jascha {Sohl-Dickstein}, Eric~A. Weiss, Niru Maheswaranathan, and Surya
  Ganguli.
\newblock Deep unsupervised learning using nonequilibrium thermodynamics.
\newblock https://arxiv.org/abs/1503.03585v8, 2015.

\bibitem{songdenoisingdiffusionimplicit_2020}
Jiaming Song, Chenlin Meng, and Stefano Ermon.
\newblock Denoising diffusion implicit models.
\newblock https://arxiv.org/abs/2010.02502v4, 2020.

\bibitem{songscorebasedgenerativemodeling_2020}
Yang Song, Jascha {Sohl-Dickstein}, Diederik~P. Kingma, Abhishek Kumar, Stefano
  Ermon, and Ben Poole.
\newblock Score-based generative modeling through stochastic differential
  equations.
\newblock https://arxiv.org/abs/2011.13456v2, 2020.

\bibitem{NEURIPS2020_634841a6}
Ximeng Sun, Rameswar Panda, Rogerio Feris, and Kate Saenko.
\newblock Adashare: Learning what to share for efficient deep multi-task
  learning.
\newblock In H. Larochelle, M. Ranzato, R. Hadsell, M.F. Balcan, and H. Lin,
  editors, {\em Advances in Neural Information Processing Systems}, volume~33,
  pages 8728--8740. Curran Associates, Inc., 2020.

\bibitem{tancik_tancikillusion-diffusion_2024}
Matthew Tancik.
\newblock tancik/{Illusion}-{Diffusion}, July 2024.
\newblock original-date: 2023-02-12T22:39:28Z.

\bibitem{vaswani2017attention}
A Vaswani.
\newblock Attention is all you need.
\newblock {\em Advances in Neural Information Processing Systems}, 2017.

\bibitem{wangpretrainingallyou_2022}
Tengfei Wang, Ting Zhang, Bo Zhang, Hao Ouyang, Dong Chen, Qifeng Chen, and
  Fang Wen.
\newblock Pretraining is all you need for image-to-image translation, 2022.

\bibitem{wangquantifyingambiguitiesartistic_2020}
Xi Wang, Zoya Bylinskii, Aaron Hertzmann, and Robert Pepperell.
\newblock Toward quantifying ambiguities in artistic images, 2020.

\bibitem{xincurrentmultitaskoptimization_2022}
Derrick Xin, Behrooz Ghorbani, Ankush Garg, Orhan Firat, and Justin Gilmer.
\newblock Do current multi-task optimization methods in deep learning even
  help?, 2022.

\bibitem{yugradientsurgerymultitask_2020}
Tianhe Yu, Saurabh Kumar, Abhishek Gupta, Sergey Levine, Karol Hausman, and
  Chelsea Finn.
\newblock Gradient surgery for multi-task learning, 2020.

\bibitem{zhang2024ctrl}
Guiyu Zhang, Huan-ang Gao, Zijian Jiang, Hao Zhao, and Zhedong Zheng.
\newblock Ctrl-u: Robust conditional image generation via uncertainty-aware
  reward modeling.
\newblock {\em arXiv preprint arXiv:2410.11236}, 2024.

\bibitem{10.1093/nsr/nwx105}
Yu Zhang and Qiang Yang.
\newblock {An overview of multi-task learning}.
\newblock {\em National Science Review}, 5(1):30--43, 09 2017.

\bibitem{zhang2021survey}
Yu Zhang and Qiang Yang.
\newblock A survey on multi-task learning.
\newblock {\em IEEE transactions on knowledge and data engineering},
  34(12):5586--5609, 2021.

\bibitem{zheng2023steps}
Yupeng Zheng, Chengliang Zhong, Pengfei Li, Huan-ang Gao, Yuhang Zheng, Bu Jin,
  Ling Wang, Hao Zhao, Guyue Zhou, Qichao Zhang, et~al.
\newblock Steps: Joint self-supervised nighttime image enhancement and depth
  estimation.
\newblock In {\em 2023 IEEE International Conference on Robotics and Automation
  (ICRA)}, pages 4916--4923. IEEE, 2023.

\end{thebibliography}
}

\newpage
\appendix

\section{Computation Efficiency}

In this section, we discuss the computational complexity of existing methods and our proposed approach for visual anagram generation.

\textbf{Experimental Setup.} We conduct the experiments on a server equipped with two AMD EPYC 7742 CPUs and 1TB RAM, running CUDA 12 and PyTorch 2.1.1, and each method only uses a single NVIDIA RTX 3090 GPU at a time. For Burgert \etal \cite{burgert_diffusion_2024} and Tancik  \cite{tancik_tancikillusion-diffusion_2024}, number of iterations and number of inference steps per image are set to 10,000 and 500, respectively, which are the default values in their open-source code. For Geng \etal \cite{geng2024visual} and our proposed method, the number of inference steps is fixed at 30. These settings are consistent with the main text. Note that fine-tuning backbone diffusion models is not required for all methods, and we report the average computation time per image for each method in \cref{tab:comp-time}. The reported time exclude model and data loading time, and all methods are evaluated using the 2-view CIFAR10 dataset as in the main text.

\textbf{Results.} As shown in \cref{tab:comp-time}, among all tested methods, Burgert \etal \cite{burgert_diffusion_2024} has the longest computation timedue to its use of Score Distillation Loss (SDL)\cite{pooledreamfusion}, which involves a large number of iterative optimization steps. The other three methods run significantly faster, as they operate within the typical time frame of diffusion model inference. Our proposed method is slightly slower than our baseline method \cite{geng2024visual}, as it incorporates additional modules to enhance visual anagram quality. Tancik \cite{tancik_tancikillusion-diffusion_2024} takes slightly more time than our method, primarily because it employs a latent diffusion model \cite{rombachhighresolutionimagesynthesis_2021} where the latent code is not rotation-invariant, and therefore requires more inference steps to generate the final image.

\begin{table}[htbp]
    \centering
    \resizebox{\linewidth}{!}
    {
        \begin{tabular}{ll}
            \toprule
            Method                                            & Computation Time (sec/img) \\
            \midrule
            Tancik\cite{tancik_tancikillusion-diffusion_2024} & 27.1                       \\
            Burgert \etal \cite{burgert_diffusion_2024}       & 3016.0                     \\
            Visual Anagrams\cite{geng2024visual}              & 13.4                       \\
            Ours                                              & 20.2                       \\
            \bottomrule
        \end{tabular}
    }
    \caption{\textbf{Computation Complexity.} We report the average computation time per image for each method.}
    \label{tab:comp-time}
\end{table}

\section{Additional Qualitative Results}

\begin{figure*}
    \centering
    \includegraphics[width=\linewidth]{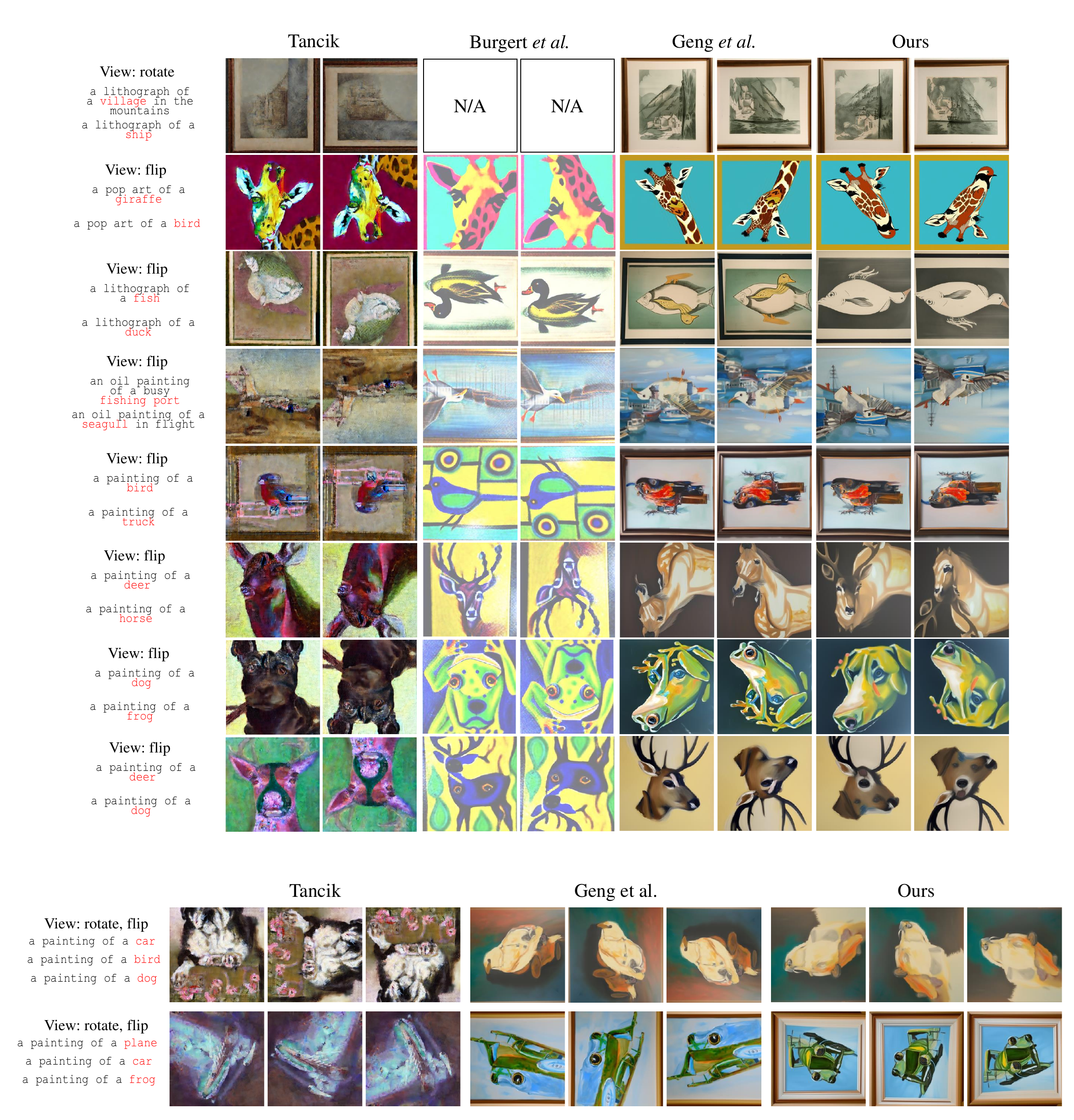}
    \caption{\textbf{Qualitative Results.} We provide additional qualitative results to compare our method with existing methods. Tancik \cite{tancik_tancikillusion-diffusion_2024} uses a latent diffusion model\cite{rombachhighresolutionimagesynthesis_2021} but struggles with transformation inconsistencies in the latent code, as discussed in \cite{geng2024visual}. Burgert \etal \cite{burgert_diffusion_2024} employs Score Distillation Loss (SDL) \cite{pooledreamfusion} which requires expensive iterative optimization and results in reduced image quality. Geng \etal \cite{geng2024visual} also encounters issues with concept segregation and concept domination. For Burgert \etal~\cite{burgert_diffusion_2024}, we only present results for 2-view flippy visual anagram generation, as their released code does not support other configurations.}
    \label{fig:supp-quali}
\end{figure*}

We provide additional qualitative results to compare our method with existing methods\cite{tancik_tancikillusion-diffusion_2024,burgert_diffusion_2024,geng2024visual} in \cref{fig:supp-quali}, including samples generated on prompts from the 2-view CIFAR10 and 3-view CIFAR10 datasets mentioned in the main text, together with some free-form examples. 

\end{document}